\pdfoutput=1

\documentclass[11pt]{article}

\usepackage[]{EMNLP2023}
\usepackage{multirow}
\usepackage{makecell}
\usepackage{arydshln} 
\usepackage{url}
\usepackage{hyperref}
\usepackage{svg}
\usepackage{tabularx}
\usepackage{booktabs}

\usepackage{longtable}

\usepackage{times}
\usepackage{latexsym}
\usepackage{amsfonts}
\usepackage[T1]{fontenc}

\usepackage[utf8]{inputenc}

\usepackage{microtype}

\usepackage{graphicx}
\usepackage{float} 
\usepackage{subfigure}
\usepackage{amsmath, xparse}
\usepackage{caption}
\usepackage{subcaption}

\title{TEQ: \underline{T}rainable \underline{E}quivalent Transformation for \underline{Q}uantization of LLMs}

\author{Wenhua Cheng \and Yiyang Cai \and Kaokao Lv \and Haihao Shen \\Intel
\\\texttt{\{wenhua.cheng, yiyang.cai, kaokao.lv, haihao.shen\}@intel.com}}
\begin{document}
\maketitle
\begin{abstract}
As large language models (LLMs) become more prevalent, there is a growing need for new and improved quantization methods that can meet the computationalast layer demands of these modern architectures while maintaining the accuracy. In this paper, we present TEQ, a trainable equivalent transformation that preserves the FP32 precision of the model output while taking advantage of low-precision quantization, especially 3 and 4 bits weight-only quantization. The training process is lightweight, requiring only 1K steps and less than $1\text{\textperthousand}$ of the original model’s trainable parameters. Furthermore, the transformation does not add any computational overhead during inference. Our results are on-par with the state-of-the-art (SOTA) methods on typical LLMs. Our approach can be combined with other methods to achieve even better performance. The code is available at https://github.com/intel/neural-compressor.
\end{abstract}

\section{Introduction}

Large language models (LLMs) have not only shown breakthrough performance in a wide range of benchmarks and tasks but played an increasingly important role in daily life, e.g., ChatGPT \citep{chatgptwebsite} in information retrieval and Copilot \citep{copilotwebsite} in programming. However, as LLMs' model size keeps growing dramatically, their significant memory footprint and heavy computation requirements become a major bottleneck of their usage. 

One of the most promising ways to alleviate this challenge is quantization, which can reduce storage and computational overhead. Quantization converts high-bit floating-point data to lower-bit representations, and it has become an effective model compression technique. 

Quantization methods can generally be divided into two categories: quantization aware training (QAT) \citep{shen2021once, zhuang2021effective, gong2019differentiable, esser2019learned, louizos2018relaxed} and post-training quantization (PTQ) \citep{frantar2022gptq, li2022efficient, xiao2022smoothquant, wei2022qdrop, frantar2022optimal, hubara2021accurate, nagel2020up, hassibi1993optimal, lecun1989optimal}. Their effectiveness has been validated for a wide range of models. However, several issues still need to be addressed, especially for LLMs. QAT simulates the quantization behavior in the training/finetuning phase, but such a process is very costly for LLMs due to their unprecedented 
parameter scale. In contrast, PTQ requires no training  and thus has drawn rising attention. However, PTQ is prone to large accuracy drops, especially for extreme low-bit quantization. This provides LLMs' PTQ methods with great opportunities for improvement.


Lower-bit quantization (e.g., Int4, W4) has recently been widely discussed since memory bandwidth is becoming the main bottleneck of LLMs. However, most existing works focus on computer vision models \citep{he2016deep, howard2017mobilenets} that are much smaller than current popular LLMs such as BLOOM-176B\citep{scao2022bloom}, OPT-175B\citep{zhang2022opt}. Other extreme quantization methods \citep{bai2020binarybert, zhang2020ternarybert} rely on the knowledge distillation technique, introducing extra overhead.  GPTQ\citep{frantar2022gptq} tunes the weights based on optimal brain surgeon\citep{hassibi1993optimal} and successfully achieves low-bit quantization on LLMs with low computation overhead.

Our proposed method reduces the compression error by introducing a trainable equivalent transformation (Fig. \ref{teq}), which keeps the mathematical equivalency of model output at FP32 precision. Moreover, the training cost is significantly low, only 1k steps of batch size 1 with around less than one-thousandth trainable parameters of the original models. Also, our method is orthogonal to current popular LLMs quantization methods, and better accuracy results could be achieved by combining ours with them. 

In summary, the contribution of this paper is threefold:
\begin{itemize}
  \item We introduce a trainable equivalent transformation for the quantization of LLMs, which keeps the model output unchanged at FP32 precision. Besides, the training is quite lightweight.
  \item Experimental results show our method could achieve results on par with or better than the SOTA methods.
  \item We also show that our method could be combined to get the new SOTA performance.
\end{itemize}

In the following, we first briefly introduce the work related to ours in Section 2. We then present the trainable equivalent transformation in Section 3. Experiments and conclusion are described in Sections 4 and 5 respectively.

\section{Related Work}

\paragraph{Quantization-aware Training.} QAT methods are widely used in model compression. By enabling finetuning process, quantized models' accuracy can often be on par with or even better than those of original models. \citep{louizos2018relaxed} introduce a differentiable quantization procedure by converting original weights and activations' distribution to categorical distributions. OQAT \citep{shen2021once} proposes a combined training scheme of architecture and quantization to acquire many quantized models. Afterward, they are converted to lower-bit models and optimized. \citep{zhuang2021effective} propose a progressive quantization scheme by quantizing activations after weights. Indeed, QAT methods are popular in relatively small-scale models, but their application in LLMs is limited due to the expensive training or even fine-tuning costs as mentioned in Section 1. 

\paragraph{Post-training Quantization.} A large number of  post-training methods quantize weights step by step and modify unquantized weights to compensate for errors produced by previously quantized weights. Optimal Brain Damage (OBD) \citep{lecun1989optimal} uses second-derivative information (Hessian-based estimation) to predict the effect of weights' perturbation analytically. Optimal Brain Surgeon (OBS) \citep{hassibi1993optimal} applies such an idea by devising a second-order framework for weight pruning. Afterward, Optimal Brain Quantization (OBQ) migrate OBS's pruning framework to quantization since pruning and quantization share the common idea of introducing perturbation in original models. Finally, GPTQ \citep{frantar2022gptq} improves the original framework's efficiency by fixing the quantization order within the layer and calculating the Hessian matrix's Cholesky decomposition before quantization. Other PTQ methods use a better rounding scheme than commonly used rounding-to-nearest (RTN). AdaRound \citep{nagel2020up} learns a rounding scheme using mean squared error (MSE) for layer-wise activation. AQuant \citep{li2022efficient} adds a learnable border function for activation quantization. 

\paragraph{Large Language Model Quantization.} Researchers are devoting efforts to compression methods particularly designed for LLMs as more open-source releases are available. LLM.int8() \citep{dettmers2022llm} discovers peak values in activation outliers' particular channels. It proposes methods to ensure that these channels are kept in higher precision. SmoothQuant \citep{xiao2022smoothquant} addresses the issues mentioned above by migrating difficulties from activation to weights with a handcrafted equivalent transformation. ZeroQuant \citep{yao2022zeroquant} devises an end-to-end quantization and inference pipeline with a novel layer-wise knowledge distillation algorithm. However, the largest model it has quantized has only 1.3B parameters. GPTQ \citep{frantar2022gptq} tunes the weights based on optimal brain surgeon \citep{hassibi1993optimal} and successfully achieves low-bit quantization on LLMs with low computation overhead. More recent, AWQ \citep{lin2023awq} propose to search the optimal scales to protect parts of weights, since they can significantly reduce the error caused by quantization. 



\section{Methodology}

\begin{figure}
\includegraphics[scale=0.47]{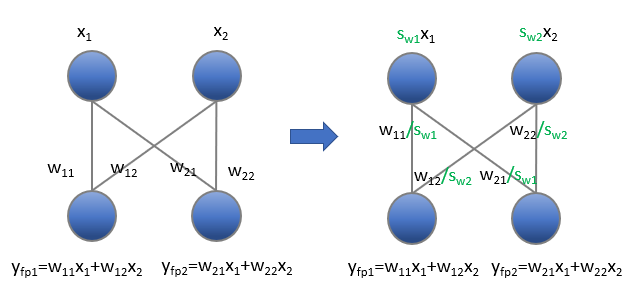}
\caption{A schematic illustration of TEQ, where $s_{w1}$ and  $s_{w2}$ are trainable parameters. A per-channel scale is multiplied at activations while an inverse scale is multiplied at weights, which could keep the output equivalent.}  
\label{teq}
\centering
\end{figure}

Figure \ref{teq} presents a schematic illustration of equivalent transformation. In the following, we introduce the quantization process first. Consider a feed-forward neural network comprised of $L$ layers, which perform matmul or convolution operations.  We only consider the matmul layer for simplicity since our method could be easily extended to convolution layers. The $l^{th}$ matmul operation can be denoted by $y_l=w_l \cdot x_l$. In which $w_l$ and $x_l$ are the weights and activation(input), and $y_l$ is the corresponding output. To quantize a tensor, a quantization op presented below could be applied.
\begin{equation}
Q(v) = clip(\left[ \frac{v}{s} \right], -n, n), n \in \mathbb{N}
\end{equation}
where $s$ denotes the quantization scale parameter and $\left[ \cdot \right]$ denotes the round-to-nearest (RTN) operation, while $-n$ and $n$ denote the integer thresholds for clipping. We ignore the zero point for simplicity.  For a  normal int8 quantization, i.e., W8A8, we need to quantize activation and weights both. And for weight-only quantization,  only the weights need to be quantized. Finally, a de-quantization operation will be appended to reconstruct the float output, normally not equal to $y_l$. In summary, the  $L_l$'s output after normal quantization  is converted to:
\begin{equation}
\hat{y_l}=Q^{-1}(Q(w_l) \cdot Q(x_l)) 
\end{equation}
where $\hat{y_l}$ denotes the $L_l$'s reconstructed output after quantization.  The value of $(y_l - \hat{y_l})^2$ is usually named as  quantization loss.

\subsection {Trainable Equivalent Transformation}
\label{sec:tet}
PTQ tends to cause a noticeable accuracy drop as mentioned before. SmoothQuant \citep{xiao2022smoothquant} and AWQ \citep{lin2023awq} rely on handcrafted rules to migrating quantization difficulties of weights and activations. However, these rules often fall into sub-optimal solutions, which cannot minimize error caused by quantization. To alleviate this issue, we introduce a trainable equivalent transformation that enforces the Fp32 output as the same but greatly improves the quantization robustness. To be more specific, suppose the shape of $w_l$ is ${c^{in}_{l}\times c^{out}_{l}}$, which stands for their respective input and output channel numbers. 
For each layer $L_l$, we can multiply a per-channel scaling vector $s_{l} \in \mathbb{R}^{c^{in}_{l}}$ for weights and append a corresponding inverse scale vector for activation. Mathematically, this can be restated as 
\begin{equation}
y_l = w_l\cdot diag(s_{l}) \cdot diag(s_{l})^{-1}\cdot x_l
\end{equation}
operator $diag(\cdot)$ denotes converting a column/row vector to a diagonal matrix whose eigenvalues are identical to the original vector's elements.
\begin{equation}
 diag \left( \begin{bmatrix} s_1 \\ s_2 \\ \vdots \\ s_n \end{bmatrix} \right)
 =
 \begin{bmatrix}
   s_{1} &  &  & \\ 
   & s_{2} &  & \\ 
   &  &  \ddots & \\ 
   &  &   & s_{n} 
 \end{bmatrix}
\end{equation}

Our observation shows the optimal $s_w$ is useful to reduce the quantization loss. Therefore, we quantize the transformed model rather than the original one. 

\begin{table*}[htbp]\small
\centering
\begin{tabular}{c | c |c |c| c| c|  c| c}
\hline
\hline
{n\_bits} & {Methods} &{OPT-6.7B} &{OPT-13B} &{BLOOM-3B} & {BLOOM-7B1} &{LLAMA-7B} & {LLAMA-13B}\\
\hline
32              & FP32  & 64.97 & 65.54 & 55.65 & 60.29 & 68.87 & 71.06\\
\hline
\multirow{4}{*}{4}    & RTN          & 62.99          & 64.17          & 53.17          & 57.80         & 67.41          & 68.86 \\
                      & GPTQ         & 63.09          & 64.83 & \textbf{54.65}          & 58.26         & 64.70         & \textbf{70.00} \\
                      & Ours         & 63.30          & 64.91          & 53.83          & 58.93         & \textbf{67.71} & 69.55\\
                      & Ours+GPTQ    & \textbf{63.94} & \textbf{65.03}          &54.42  & \textbf{59.62}& 65.27          & 69.73 \\

\hline
\multirow{4}{*}{4\_g128}& RTN        & 64.04          & 64.88          & 54.91          & 59.32         & 67.87          & 70.88 \\
                        & GPTQ       & 64.76 & \textbf{65.37} & \textbf{55.68} & 59.59         & 66.33         & 70.92\\
                        & Ours       & 64.11          & 64.87          & 54.98          & 59.35         &\textbf{68.10}  & \textbf{71.00}\\
                        & Ours+GPTQ  & \textbf{64.77}          & 65.20          & 55.49          &\textbf{59.60} & 66.56          & 70.96 \\
                     
\hline
\hline
\end{tabular}
\caption{The w4 average accuracy($\uparrow$) of four tasks, e.g., HellaSwag, WinoGrande, PIQA, and LAMBADA, in LM-eval. g denotes group size. "Ours+GPTQ" means we apply TEQ first and then apply GPTQ afterward. For LLAMA-7B, the result of GPTQ is w/o act-order.  Results of act-order are shown in Appendix \ref{A.1}.}
\label{table1}
\end{table*}

\begin{table*}[htbp]\small
\centering
\begin{tabular}{c | c |c |c| c| c|  c| c}
\hline
\hline
{n\_bits} & {Methods} &{OPT-6.7B} &{OPT-13B} &{BLOOM-3B} & {BLOOM-7B1} &{LLAMA-7B} & {LLAMA-13B}\\
\hline
32             & FP32  & 10.86 & 10.12 & 13.48 & 11.36 & 5.68 & 5.09 \\
\hline
\multirow{4}{*}{4 }  & RTN       & 12.10          & 11.32          & 14.75          & 12.09          & \textbf{6.29} & 5.53 \\
                     & GPTQ      & 11.59          & \textbf{10.33} & 14.10          & \textbf{11.73} & 6.59          & \textbf{5.33} \\
                     & Ours      & 11.68          & 10.59          & 14.72          & 12.21          & 6.30         & 5.50 \\
                     & Ours+GPTQ & \textbf{11.29} & 10.36          & \textbf{14.03} & 11.74          & 6.76        & 5.35 \\

\hline
\multirow{4}{*}{4\_g128}& RTN       & 11.16          & 10.32          & 13.85          & 11.60          & \textbf{5.97} & 5.26 \\
                        & GPTQ      & \textbf{10.98} & \textbf{10.20} & \textbf{13.69} & \textbf{11.48} & 6.29          & \textbf{5.21} \\
                        & Ours      & 11.11          & 10.28          & 13.82          & 11.58          & \textbf{5.97} & 5.26 \\
                        & Ours+GPTQ & 11.02 & 10.21          & \textbf{13.69} & \textbf{11.48} & 6.28          & \textbf{5.21} \\
                     
\hline
\hline
\end{tabular}
\caption{The w4 perplexity($\downarrow$) on WikiText-2. For LLAMA-7B, the result of GPTQ is w/o act-order. Results of act-order are shown in Appendix \ref{A.1}.}
\label{table2}
\end{table*}

The transformation has two per-channel scale operations, which will introduce computation overhead. We fuse the weight scale to  the weight itself. For the activation scale, following \citep{xiao2022smoothquant}, we fuse it to the previous layers, such as layernorm\citep{ba2016layer}, batchnorm\citep{ioffe2015batch} and etc. In all our experiments, we only apply the transformation to the layer whose scales could be fused, which introduces no extra overhead at deployment. 

\subsection {Training Details} \label{sec3.2}
We train the scales $s_l$ because there is little knowledge of the best equivalent transformation due to various models and quantization configurations. It's worth mentioning that the count of trainable scales is much less than the model's parameters, and the model weights are frozen.

To train the transformation scales, we follow the basic QAT to simulate the quantization behavior, which could be denoted as  

\begin{equation}
y_{l_q}=(Q^{-1}Q(w_l))(Q^{-1}Q(x_l))
\end{equation}

For weight-only quantization, activation quantization will be ignored. We adopt straight-through estimator (STE) \citep{bengio2013estimating} to backward the  gradients.
 
 We use Adam\citep{kingma2014adam} optimizer, betas [0.9, 0.9], and weight decay  0. The learning rate is 1e-3 unless explicitly stated and the decay type is linear.  We only train 1000 steps. We use the same loss function as the original one in the training phase. For example, CrossEntorpy loss is adopted for LLMs. The $s_l$ is usually initialized with 1. However, sometimes $1.0/sqrt(w_{cin})$ leads to better results, so we pick the better one in our experiments.

\section{Experiments}
In this section, we evaluate our proposed TEQ's in different aspects. Initially, we briefly introduce LLM architectures and tasks included in our evaluation. Secondly, we illustrate a detailed comparison of our method and other state-of-the-art (SOTA) methods, and both quantization accuracy and time are considered. 

\subsection{Experimental Settings}

\paragraph{Large Language Models.}
We conduct our experiments on the most popular LLM architectures, including LLaMAs \citep{touvron2023llama}, BLOOMs \citep{scao2022bloom}, and OPTs \citep{zhang2022opt}. Parameter scalings ranging from million to billion are all included.  

\begin{table*}[htbp]\small
\centering
\begin{tabular}{c | c |c |c| c| c|  c| c}
\hline
\hline
{n\_bits} & {Methods} &{OPT-6.7B} &{OPT-13B} &{BLOOM-3B} & {BLOOM-7B1} &{LLAMA-7B} & {LLAMA-13B}\\
\hline
32              & FP32  & 64.97 & 65.54 & 55.65 & 60.29 & 68.87 & 71.06\\
\hline
\multirow{4}{*}{3\_g128}& RTN   & 56.03 & 49.59 & 52.54 & 57.53 & 64.92 & 67.68 \\
                        & GPTQ  & 62.98 & \textbf{64.68} & 53.41 & \textbf{58.12} & 58.29 & 68.73 \\
                        & Ours  & 61.41 & 63.27 &  52.69& 57.79 & \textbf{65.25} & 68.32 \\
                         & Ours+GPTQ  & \textbf{63.16} & 64.60 & \textbf{53.71} & 58.00 & 59.27 &  \textbf{69.15}\\
\hline
\hline
\end{tabular}
\caption{The 3 bits with group size 128 average accuracy($\uparrow$) of four tasks,e.g., HellaSwag, WinoGrande, PIQA, and LAMBADA, in LM-eval. g denotes group size. For LLAMA-7B, the result of GPTQ is w/o act-order. Results of act-order are shown in Appendix \ref{A.1}.}
\label{table3}
\end{table*}

\begin{table*}[htbp]\small
\centering
\begin{tabular}{c | c |c |c| c| c|  c| c}
\hline
\hline
{n\_bits} & {Methods} &{OPT-6.7B} &{OPT-13B} &{BLOOM-3B} & {BLOOM-7B1} &{LLAMA-7B} & {LLAMA-13B}\\
\hline
32             & FP32  & 10.86 & 10.12 & 13.48 & 11.36 & 5.68 & 5.09 \\
\hline
\multirow{4}{*}{3\_g128}& RTN   & 22.37 & 40.50 & 15.68 & 12.47 & 7.01 & 5.88 \\
                        & GPTQ  & 11.42 & \textbf{10.51} & 14.67 & 11.99& 8.28 & \textbf{5.64} \\
                        & Ours  & 12.03 & 11.83 & 15.48 & 12.40 & \textbf{6.89} & 5.81 \\
                        & Ours+GPTQ  & \textbf{11.40} & 10.52 & \textbf{14.64} & \textbf{11.98} & 7.71 & \textbf{5.64} \\
\hline
\hline
\end{tabular}
\caption{WikiText-2 perplexity($\downarrow$) of 3 bits with group size 128. For LLAMA-7B, the result of GPTQ is w/o act-order. Results of act-order are shown in Appendix \ref{A.1}.}
\label{table4}
\end{table*}

\paragraph{Evaluation and Datasets.}
We make assessments on several language tasks to satisfy the task-agnostic setting. Specifically, we report average accuracy result on four common sense reasoning tasks by leveraging lm-eval-harness\citep{eval-harness}, including HellaSwag \citep{zellers2019hellaswag}, WinoGrande \citep{sakaguchi2021winogrande}, PIQA \citep{bisk2020piqa} and LAMBADA \citep{paperno2016lambada}. Furthermore, we complement our evaluation with perplexity (PPL) analysis on WikiText2 \citep{merity2016pointer}, PTB \citep{marcus1994penn} as well as C4 \citep{raffel2020exploring}. 

\paragraph{Implementation Details.}
Following GPTQ \citep{frantar2022gptq},  we focus on weight-only quantization and exclude the last layer When quantifying. We used a single HW accelerator to quantize models with a scale of around ten billion parameters. We use the same calibration dataset pile-10k\footnote{https://huggingface.co/datasets/NeelNanda/pile-10k} for a fair comparison. 

\paragraph{Baseline.}
Our primary baseline is vanilla round-to-nearest quantization (RTN) which has a remarkable result at 4bits using a small group size of 128. We also compare with a state-of-the-art method GPTQ \citep{frantar2022gptq}. 

\subsection{Results}
 As mentioned above, we compare our results with RTN and the SOTA GTPQ\citep{frantar2022gptq}. Also, since our method is orthogonal to GPTQ, we report Ours+GPTQ as well, which applies TEQ first and then runs GPTQ official code\footnote{https://github.com/IST-DASLab/gptq} afterward. We mainly focus on the models around 10B which is commonly used.
\paragraph{W4 Quantization.}
We first evaluate TEQ on popular 4 bits quantization. Table \ref{table1} shows the lm-eval results of different LLM model architectures and parameter sizes. TEQ outperforms RTN in all cases except one. Comparing with GPTQ, TEQ shows better results in 6 out of 12 scenarios. After combining GPTQ, new state-of-the-art results could be achieved in 5 scenarios. In summary, TEQ could be helpful in 8 out of 12 scenarios. Table \ref{table7} shows the hyper-parameters that we used in the experiements.

 We also evaluate WikiText2 ppl in table \ref{table2} w/o group size and group size 128. TEQ is better or on par with RTN. Similarly, the combined approach (Ours and GPTQ) shows comparable or better results than standalone GPTQ.

\paragraph{W3 Quantization.} We also evaluate TEQ at weight with 3 bits. We only consider group size 128, because the performance drops a lot without group size and usually could not be deployed in practice. Similar to 4 bits evaluation, we report the lm-eval result and wikitext2 ppl result in table \ref{table3} and \ref{table4} respectively. TEQ outperforms RTN in all scenarios and is inferior to GPTQ on certain models. However, TEQ could bring improvement for 8 out of 12 scenarios if taking Ours+GPTQ into account.

\paragraph{Quantization Time.}
We report the quantization time in Table \ref{table5}. We adopt Deepspeed\footnote{https://github.com/microsoft/DeepSpeed} for 10B+ models due to the potential out-of-memory (OOM) issue. As TEQ needs training, our time cost is reasonably higher than GPTQ, especially when the model does not fit into the device memory. It's possible to reduce the time further by using more resources or optimizing the code, while it's out of scope.

\begin{table}[htbp]\small
\centering
\begin{tabular}{ c |c |c  }
\hline
\hline
 {Models} & {GPTQ} & {Ours} \\
 \hline
  OPT-6.7B  & 841  & 1239  \\
\hline
  OPT-13B   & 1523 & 8737* \\      
   \hline
BLOOM-3B  & 345  & 506  \\
\hline
   BLOOM-7B1 & 661 &  1148 \\  
 \hline
   LLAMA-7B  & 712  & 1249  \\
\hline
 LLAMA-13B & 1240 & 9501*  \\  
\hline
\hline
\end{tabular}
\caption{Quantization time in seconds for 4-bit weight quantization. * denotes DeepSpeed is adopted in training for 10B+ models.}
\label{table5}
\end{table}

\begin{figure*}[!t]
\centering
\includegraphics[scale=1.07]{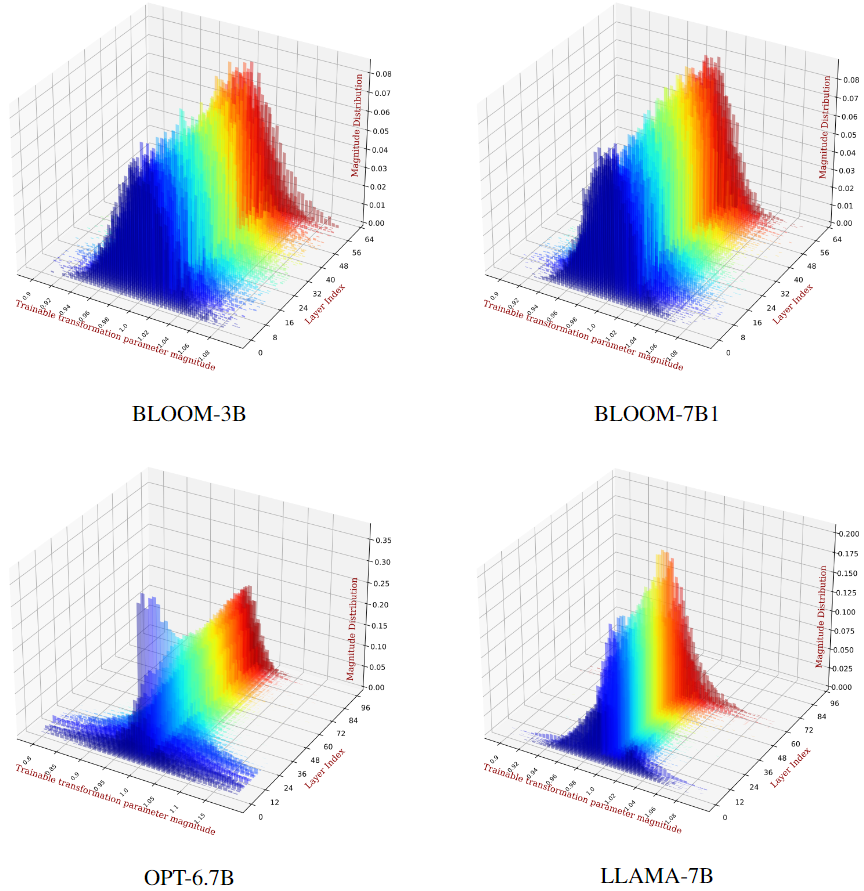}
\caption{ The magnitude distributions of scales in TEQ for \textbf{BLOOM-3B}, \textbf{BLOOM-7.1B},  \textbf{OPT-6.7B}, \textbf{LLAMA-7B}. The quantization configurations are w3\_g128, w4\_g128, w4, and w4 respectively. Different colors refer to layer indices in models (blue stands for shallow layers which are close to the data layer,  while red stands for deeper layers).}
\label{vis_new1}
\end{figure*}

\paragraph{Analysis of Scales in TEQ.} \label{vis:mag}  We visualize the magnitude distribution histograms of $s_l$ for the layers to which TEQ can be applied. 
Figure \ref{vis_new1} displays the results of models with $s_l$ initialized as scalar ones. Several conclusions can be drawn from these results. Most notably, the majority of trained scales remain close to their initial values (e.g., 1), typically within the range of [0.75, 1.25]. This suggests that even minor changes to the model can significantly reduce quantization loss. Additionally, some scales deviate considerably from 1, indicating the presence of “outlier” channels. Furthermore, scales in middle layers tend to remain closer to their initial values compared to other layers, suggesting that the first and last layers are more sensitive to the quantization loss. We also attach results of scales initialized with $1.0/sqrt(w_{cin})$ in Appendix \ref{A.4}.

\section{Conclusion}
In this paper, we propose TEQ, a trainable equivalent transformation that preserves the FP32 precision of the model output while also taking advantage of low-precision quantization, and its training process is lightweight. Plus, TEQ is regarded as orthogonal support for other quantization methods to improve their performance. Our task-agnostic experiments and comparison with other methods show that TEQ or its combination with other methods can obtain comparable or better results. 

\subsection{Limitations}
We find that the required memory during training is still high, though the number of training parameters remains low. Moreover, since we enforce the transformation to be equivalent and keep the architecture and FP32 output unchanged, our results in some scenarios are inferior to the SOTA methods, which could be fixed by combining the SOTA methods.

\subsection{Ethics Statement}
We propose TEQ for LLMs quantization. The method can be either used individually or combined with other quantization methods. Since TEQ only requires a few steps of finetuning on original models. Thus, it is safe to say that TEQ's technical details have no significant ethical implications. Our work provides an exploration of large language model quantization through simple finetuning, making their application easier. We believe increasingly more work like this will emerge, making LLMs' quantization more powerful. 

\bibliography{anthology,custom}
\bibliographystyle{acl_natbib}

\appendix

\section{Appendix}
\label{sec:appendix}
\newcommand{\tabincell}[2]{\begin{tabular}{@{}#1@{}}#2\end{tabular}} 

\subsection{Additional comparison with AWQ}
\label{A.2}
Although both AWQ and TEQ use a small calibration set from Pile, TEQ's evaluation methodology closely follows that of GPTQ and only shares a few common tasks with AWQ. It is important to acknowledge that this comparison inherently lacks rigor due to our reliance on referencing AWQ's data alone. Consequently, this approach introduces the potential unfairness in the evaluation process, primarily stemming from the utilization of different calibration datasets.

\begin{table*}[t]
\begin{center}
\label{table awq}
\begin{tabular}{c|c|ccc|ccc}
\hline
\hline
 \multicolumn{2}{c|}{LLaMA-7B}   & \multicolumn{3}{c|}{AWQ} & \multicolumn{3}{c}{Ours} \\ \hline 

 \multirow{1}{*}{nbits}&   \multirow{1}{*}{Method} & \multicolumn{1}{c}{PIQA} & \multicolumn{1}{c}{Hella.} & \multicolumn{1}{c|}{Wino.}  & \multicolumn{1}{c}{PIQA} & \multicolumn{1}{c}{Hella.} & \multicolumn{1}{c}{Wino.} \\ \hline
    
  \multirow{1}{*}{16}& \multicolumn{1}{c|}{FP16} & \multicolumn{1}{c}{78.35} & \multicolumn{1}{c}{56.44} & \multicolumn{1}{c|}{67.09} &\multicolumn{1}{c}{78.35}  & \multicolumn{1}{c}{56.42} & \multicolumn{1}{c}{66.85} \\ \hline

 \multirow{3}{*}{W3G128}& \multicolumn{1}{c|}{RTN} & \multicolumn{1}{c}{75.84} & \multicolumn{1}{c}{53.10} & \multicolumn{1}{c|}{63.22} &\multicolumn{1}{c}{75.68}  & \multicolumn{1}{c}{53.18} & \multicolumn{1}{c}{\textbf{63.06}} \\

& \multicolumn{1}{c|}{GPTQ} & \multicolumn{1}{c}{70.89} & \multicolumn{1}{c}{46.77} & \multicolumn{1}{c|}{60.93} &\multicolumn{1}{c}{72.58}  & \multicolumn{1}{c}{47.10} & \multicolumn{1}{c}{59.91} \\

& \multicolumn{1}{c|}{Proposed} & \multicolumn{1}{c}{\textbf{76.66}} & \multicolumn{1}{c}{\textbf{53.63}} & \multicolumn{1}{c|}{\textbf{66.14}} &\multicolumn{1}{c}{\textbf{76.01}}  & \multicolumn{1}{c}{\textbf{53.30}} & \multicolumn{1}{c}{\textbf{63.06}} \\ \hline

 \multirow{3}{*}{W4G128}& \multicolumn{1}{c|}{RTN} & \multicolumn{1}{c}{77.86} & \multicolumn{1}{c}{\textbf{55.81}} & \multicolumn{1}{c|}{65.59} &\multicolumn{1}{c}{77.58}  & \multicolumn{1}{c}{\textbf{55.91}} & \multicolumn{1}{c}{65.59} \\

  & \multicolumn{1}{c|}{GPTQ} & \multicolumn{1}{c}{77.20} & \multicolumn{1}{c}{53.98} & \multicolumn{1}{c|}{65.67} &\multicolumn{1}{c}{77.58}  & \multicolumn{1}{c}{55.83} & \multicolumn{1}{c}{\textbf{66.54}} \\

 & \multicolumn{1}{c|}{Proposed} & \multicolumn{1}{c}{\textbf{78.07}} & \multicolumn{1}{c}{55.76} & \multicolumn{1}{c|}{\textbf{65.82}} &\multicolumn{1}{c}{\textbf{78.02}}  & \multicolumn{1}{c}{55.76} & \multicolumn{1}{c}{\textbf{66.54}} \\ 
\hline
\hline
\end{tabular}
\end{center}
\caption{Reported results of AWQ and Ours} 
\end{table*}

Table 6 presents the LLaMA-7B's results of our common tasks alongside AWQ in table below and all the results of AWQ are from their paper.


\subsection{Additional comparison with GPTQ act-order}
\label{A.1}
We show the results in Table \ref{table 6}. TEQ still outperforms GPTQ in most cases.

\begin{table}[H]\small
\centering
\begin{tabular}{c | c |c |c}
\hline
\hline
{nbits / gs} & {Methods} &{lm-eval ($\uparrow$)} & \makecell{wikitext2 \\ ppl ($\downarrow$)} \\
\hline
\multirow{3}{*}{4 / -1}& GPTQ-AO  &  0.6713 & 6.06 \\
                        & Ours  & \textbf{0.6771} &  6.30 \\
                        & Ours+GPTQ-AO  & 0.6736 & \textbf{6.03} \\
\hline
\multirow{3}{*}{4 / 128}& GPTQ-AO  &  0.6809 & 5.82 \\
                        & Ours  & \textbf{0.6813} &  5.97 \\
                        & Ours+GPTQ-AO  & 0.6811 & 5.82 \\
\hline
\multirow{3}{*}{3 / 128}& GPTQ-AO  &  0.6042 & 8.29 \\
                        & Ours  & 0.6521 &  6.89 \\
                        & Ours+GPTQ-AO  & \textbf{0.6647} & \textbf{6.61} \\
\hline
\hline
\end{tabular}
\caption{Comparing results of Llama-7B for GPTQ with act-order. AO denotes act-order. TEQ still outperforms GPTQ in most cases.}
\label{table 6}
\end{table}

\subsection{Special hyperparameters and settings}
Usually, we adopt the same hyperparameters mentioned in section \ref{sec3.2}. So we only list all the particular settings in Table \ref{table7}.

\begin{table}[H]\small
\centering
\begin{tabular}{ c |c |c}
\hline
\hline
{lr} &{initialization} & {Models}\\
\hline
default & $1.0/sqrt(w_{cin})$ & \tabincell{c}{opt13b\_w4; bloom3b\_w4;\\bloom7b\_w4;\\opt6.7b\_w4\_g128;\\opt13b\_w3\_g128}  \\
\hline
4e-4 &  default & llama7b\_w4;   \\
\hline
2e-4 &  default & llama13b\_w4\_g128;   \\
                         
\hline
\hline
\end{tabular}
\caption{Special hyperparameters and settings. g denotes group size}
\label{table7}
\end{table}

\begin{figure*}[!t]
\centering
\includegraphics[scale=0.85]{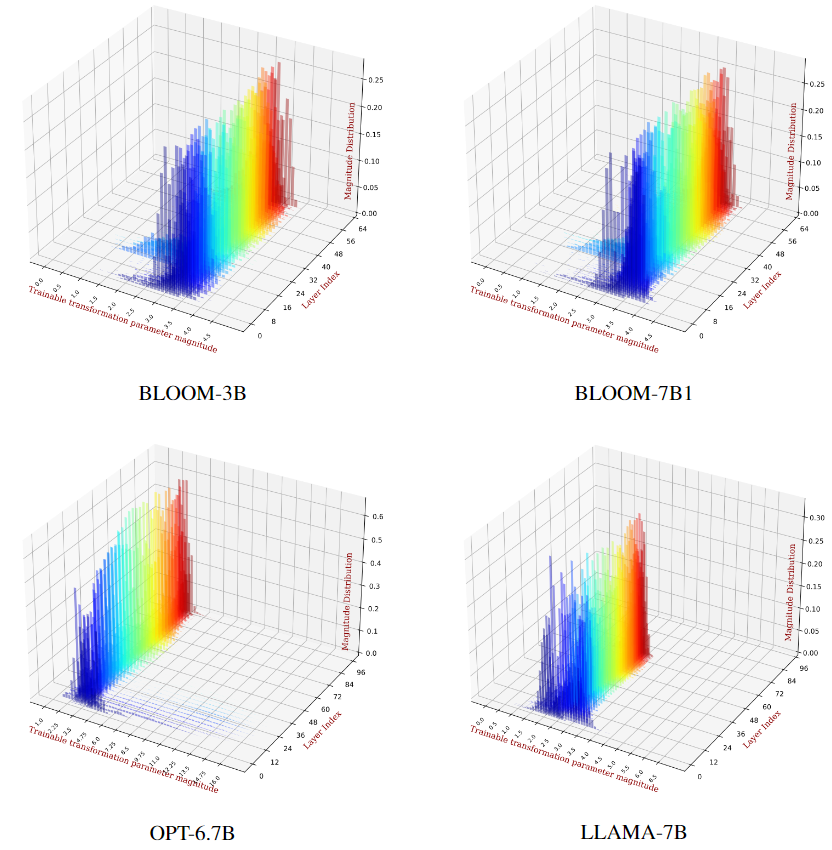}
\caption{TEQ's trained transformation parameters' magnitude distributions, using maximum's square root value for initialization. From top to down are BLOOM-3B, BLOOM-7.1B, OPT-6.7B and LLAMA-7B respectively.}
\label{vis_new2}
\end{figure*} 

\subsection{More visualization results for TEQ's trained parameters.}
\label{A.4}
Figure \ref{vis_new2}  shows  the magnitude distribution of scales initialized with $1.0/sqrt(w_{cin})$. Since the initial value is related to channel-wise maximum values, it's more challenging to analyze. However, some outliers could be still observed.

\subsection{Counts of trainable parameters introduced by TEQ}
\label{A.4}
We provide more details about counts of trainable parameters introduced by TEQ in Table 9. table presented below offers details regarding the applicable layers of TEQ in several models. We handle linear layers that possess transformation scales that can be assimilated by their preceding layers, such as Layer Normalization, among others.

As an illustration, within a single transformer block of OPT-6.7B, the QKV layers have the same preceding layers and therefore utilize the same set of trainable parameters. Based on the statistics, we have observed that TEQ’s training only requires a minimal number of parameters (around the order from 1e-5 to 1e-4), thereby making our approach light-weighted enough.

\begin{table*}
\centering
\label{tab:1}       
\begin{tabular}{c|c|c|c|c|c|c|c}
    \hline
    \hline
    Models & \makecell{Blocks} & \makecell{TEQ \\ Applicable \\ Linear \\ Layers} & \makecell{Total \\ Linear \\ Layers} & \makecell{TEQ \\ Parameter \\ Groups} & \makecell{TEQ \\ Parameters \\ Counts} & \makecell{Total \\ Parameters \\ Counts} & \makecell{Ratio \\ TEQ params and \\Total Params} \\
    \hline
    \makecell{Bloom\\3B} & 30  &  60 & 121 & 60 & 153600 & 3644810240 & 0.00421\% \\
    \hline
    \makecell{Bloom\\7B1} & 30 &  60  & 121 & 60 & 245760 & 8096620544 & 0.00304\% \\
    \hline
    \makecell{OPT\\6.7B} & 32 & 160  & 193 & 72 & 786432 & 6864388096 & 0.01146\% \\
    \hline
    \makecell{OPT\\13B} & 40 & 200  & 241 & 96 & 1228800 & 13110865920 & 0.00937\% \\
    \hline
    \makecell{Llama\\7B} & 32 & 160  & 225 & 64 & 262144 & 6738415616 & 0.00389\% \\
    \hline
    \makecell{Llama\\13B} & 40 & 200 & 281 & 80 & 409600 & 13015864320 & 0.00315\% \\
    \hline
    \hline
\end{tabular}
\caption{Analysis of TEQ Parameters. TEQ only require a minimal ratio of original models' parameters (around the order from 1e-5 to 1e-4).}
\end{table*}

\end{document}